\documentclass[journal]{IEEEtran}
\usepackage[utf8]{inputenc}
\usepackage{cite}
\usepackage{hyperref}
\usepackage{multirow}
\usepackage{algorithmic}
\usepackage{orcidlink}
\usepackage{subfigure}
\usepackage{amsmath} 
\usepackage{amsfonts,amssymb}  
\usepackage{colortbl}
\usepackage{algorithm,algorithmic}
\graphicspath{ {./figures/} }
\renewcommand\eqref[1]{(\autoref{#1})}
\hypersetup{
	colorlinks=true,
	linkcolor=cyan,
	filecolor=blue,      
	urlcolor=black,
	citecolor=green,
}

\begin{document}

\title{White-Box \textit{m}HC: Electromagnetic Spectrum–Aware and Interpretable Stream Interactions for Hyperspectral Image Classification}

\author{
Yimin Zhu, Lincoln Linlin Xu, ~\IEEEmembership{Member,~IEEE}, Zhengsen Xu, Zack Dewis, Mabel Heffring, Saeid Taleghanidoozdoozan, Motasem Alkayid, Quinn Ledingham, Megan Greenwood % stops a space

% \thanks{Corresponding author Lincoln Linlin Xu is with the Department of Geomatics
% Engineering, University of Calgary, Canada (email: lincoln.xu@ucalgary.ca)}

\thanks{Authors are all from the Department of Geomatics Engineering, University of Calgary, Canada,  email: yimin.zhu@ucalgary.ca}}

\markboth{Journal of \LaTeX\ Class Files,~Vol.~13, No.~9, September~2014}
{Shell \MakeLowercase{\textit{et al.}}: }
\maketitle

\begin{abstract}

In hyperspectral image classification (HSIC), most deep learning models rely on opaque spectral–spatial feature mixing, limiting their interpretability and hindering understanding of internal decision mechanisms. We present physical spectrum-aware white-box \textit{m}HC, named ES-\textit{m}HC, a hyper-connection framework that explicitly models interactions among different electromagnetic spectrum groupings (residual stream in \textit{m}HC) interactions using structured, directional matrices. By separating feature representation from interaction structure, ES-\textit{m}HC promotes electromagnetic spectrum grouping specialization, reduces redundancy, and exposes internal information flow that can be directly visualized and spatially analyzed. Using hyperspectral image classification as a representative testbed, we demonstrate that the learned hyper-connection matrices exhibit coherent spatial patterns and asymmetric interaction behaviors, providing mechanistic insight into the model’s internal dynamics. Furthermore, we find that increasing the expansion rate accelerates the emergence of structured interaction patterns. These results suggest that ES-\textit{m}HC transforms HSIC from a purely black-box prediction task into a structurally transparent, partially white-box learning process.

\end{abstract}

% Note that keywords are not normally used for peerreview papers.

\IEEEpeerreviewmaketitle

\section{Introduction}

Hyperspectral image (HSI) classification is a fundamental task that transforms raw HSI data into valuable maps that support various key environmental and resource exploitation
tasks. Nevertheless, efficient HSI classification is challenging due to various difficult HSI characteristics, e.g., high-dimensionality, noise, Spectral-Spatial heterogeneity, and limited training samples. Given these difficulties, it is challenging to extract discriminative features that can efficiently capture subtle differences among HSI classes.

Various approaches have been proposed for dimension reduction to deal with the high-dimensionality challenge. For example, principal component analysis (PCA) \cite{rodarmel2002principal} and independent component analysis (ICA) \cite{stone2002independent} have been used to extract compact spectral features from HSI. The deep learning-based (DL) approaches, including patch-based and patch-free methods, are designed by adding more layers and pooling layers to expand the receptive field. While transitional mechanism learning struggles with capturing the non-linear feature from the hyperspectral image, boundary preservation is a hard trade-off with high accuracy in DL methods \cite{SSRN, mei2024novel}. Most of the DL models separate two branches for spatial and spectral feature extraction to deal with the ambiguity and heterogeneity. For example, Transformers are used to model the long-distance spatial context dependency \cite{9565208, SSFTT}. The limitation of the self-attention mechanism in Transformers is that computational complexity grows quadratically with respect to the size of the image (or sequence length). Compared with Transformers, the Mamba model adopts state recursion
and sequential tokens, leading to linear complexity, which reduces computations while maintaining the long-range modeling capacity. However, the traditional vision Mamba model \cite{zhu2024vision} uses a predefined sequence scanning mechanism, but lacks token sparsity, and can not choose and permute the tokens dynamically \cite{wang2024graph}. By stacking multiple layers in HSIC models, leading to underlying overfitting problems, considering the limited training samples.

The spectrum-aware grouping method is used in \cite{wang2023dcn}, but this method only randomly selects three bands to form Tri-spectral datasets, leading to less physical meaning in the electromagnetic spectrum grouping. Additionally, the pretrained model on visual images is used for feature learning, but it lacks the capability of feature representation. There are some techniques to explain the DL model, for example Gradient map (Grad-CAM) and the self-attention matrix, but Grad-CAM is a post-hoc explanation. Furthermore, although the self-attention matrix is learned from data and features, this symmetric attention matrix is still difficult to interpret physically, and often tells the token-to-token relationship, instead of the large-scale level. 

Recent advances in deep learning architectures have explored expanding the width of residual streams to enhance model capacity through dense connectivity and multi-path structures \cite{xie2026mhc, mishra2026mhc}. Hyper-Connections (HC) \cite{zhu2024hyper} use learnable matrices to build connections between different residual streams. However, the unconstrained nature of HC compromises the identity mapping property when the architecture extends across multiple layers, leading to gradient explosion \cite{xie2026mhc}. Very recently, the DeepSeek group \cite{xie2026mhc} proposed Manifold-Constrained Hyper-Connections (\textit{m}HC) for language models, addressing the gradient vanishing and explosion issue by projecting connection matrices \(\mathcal{H}^{\text{res}}\) onto the Birkhoff polytope of doubly stochastic
matrices via Sinkhorn-Knopp normalization and restoring the identity mapping
property. The mechanism of \textit{m}HC enables shallower networks to achieve comparable capacity by performing more diver transitions and wider parallel residual streams. However, there is no exploration using \textit{m}HC for the image classification task, and the \textit{m}HC didn't visualize the doubly stochastic
matrices to explain the interaction among each residual stream. This paper is the first step in analyzing the doubly stochastic matrices, forming interpretable residual stream interactions.

Hence, based on the macro design of \textit{m}HC, we introduce \textit{m}HC into HSIC task, with following contribution:

1) This is the first paper that uses \textit{m}HC for hyperspectral image classification.

2) Electromagnetic spectrum–aware residual stream approach is proposed. Instead of duplicating the feature like HC and \textit{m}HC did, we fully take advantage of the physical and meaningful spectrum characteristic in hyperspectral image, we divide the hyperspectral cube into four electromagnetic spectra groups, i.e., visible light (VIS, 400-700 nm), near-infrared (NIR, 700-1000 nm), and
shortwave infrared 1 (SWIR1, 1000-1800 nm) and shortwave infrared 2 (SWIR2, 1800-2500 nm), according to their wavelength, forming residual streams with physical significance.

3) We visualize and analyze the doubly stochastic matrices and demonstrate
that the learned hyper-connection matrices exhibit coherent spatial patterns and asymmetric interaction behaviors, providing mechanistic insight into the model’s internal dynamics, making the proposed ES-\textit{m}HC, towards the partially white-box model.

The following section will focus on methodology \autoref{method}, experimental results \autoref{exp}, and discussion and conclusion \autoref{discussion}.
% \section{Background}
\begin{figure*}[!h]
    \centering
    \includegraphics[width=\textwidth]{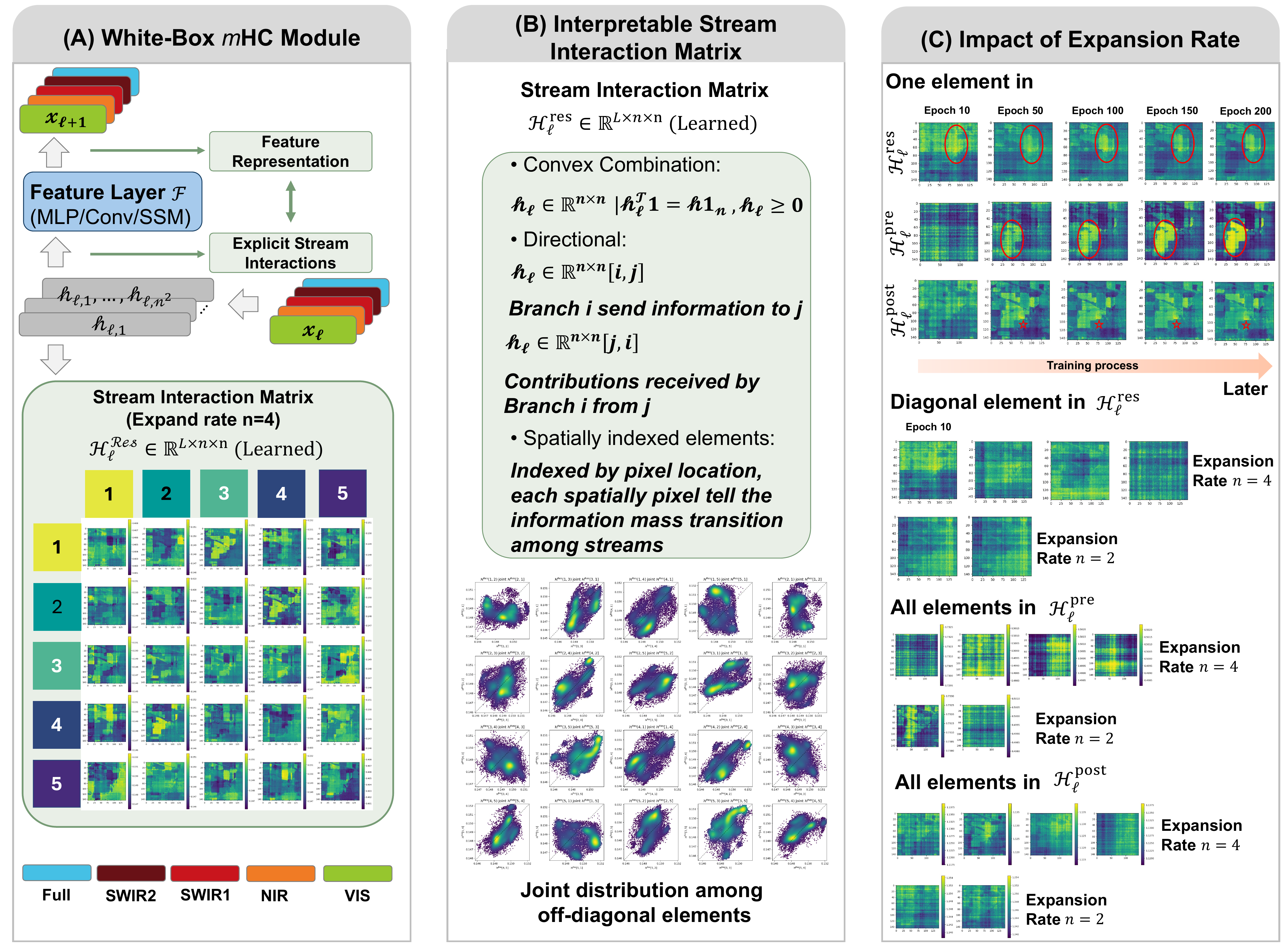}
    \caption{Illustration of the (A) model overview, (B) stream matrix, and (C) the impact of the expansion rate on the emergence of spatial pattern.}
    \label{fig1}
\end{figure*}
\section{Methodology} \label{method}

We keep the macro design of \textit{m}HC, with the following micro design specifically for the hyperspectral and remote sensing field:

\begin{itemize}
    \item \textbf{Electromagnetic spectrum–aware residual stream}: Instead of duplicating the feature, we perform wavelength-aware residual stream expansion to broaden the width of the residual stream, increasing diver feature representation at the same model layer. Four electromagnetic spectra groups are built, including VIS, NIR, SWIR1, and SWIR2, i.e., expansion rate \(n = 5\), including the full bands. As shown in \autoref{fig1} (A).
    \item \textbf{Cluster-wise sequence scanning}: We found that there is a clear spatial clustering effect in the doubly stochastic matrix, \(\mathcal{H}^{\text{res}} \in \mathbb{R}^{L \times n \times n}, L=H \times W\), as shown in \autoref{fig1} (B), and during model training, this spatial clustering pattern remains nearly consistent for each element in \(\mathcal{H}^{\text{res}} (i, j) \in \mathbb{R}^{H \times W}, 1 \leq i \leq n,  1 \leq j \leq n\). Inspired by this, we introduced cluster-wise sequence scanning for the Mamba model, where only limited tokens are selected, which can relieve the hidden problem brought by the very long sequence. 
    \item \textbf{Spectral-Spatial Mamba Block}: The selection of feature layer function \(\mathcal{F}\) in \textit{m}HC arbitrary. Considering the Spectral-Spatial heterogeneity of the hyperspectral image, the Spectral-Spatial Mamba block is designed and used as the layer function \(\mathcal{F}\), where the spatial Mamba is cluster-wise, and the spectral Mamba is used for spectral feature modeling.
\end{itemize}

\begin{figure}[!h]
    \centering
    \includegraphics[width=0.49\textwidth]{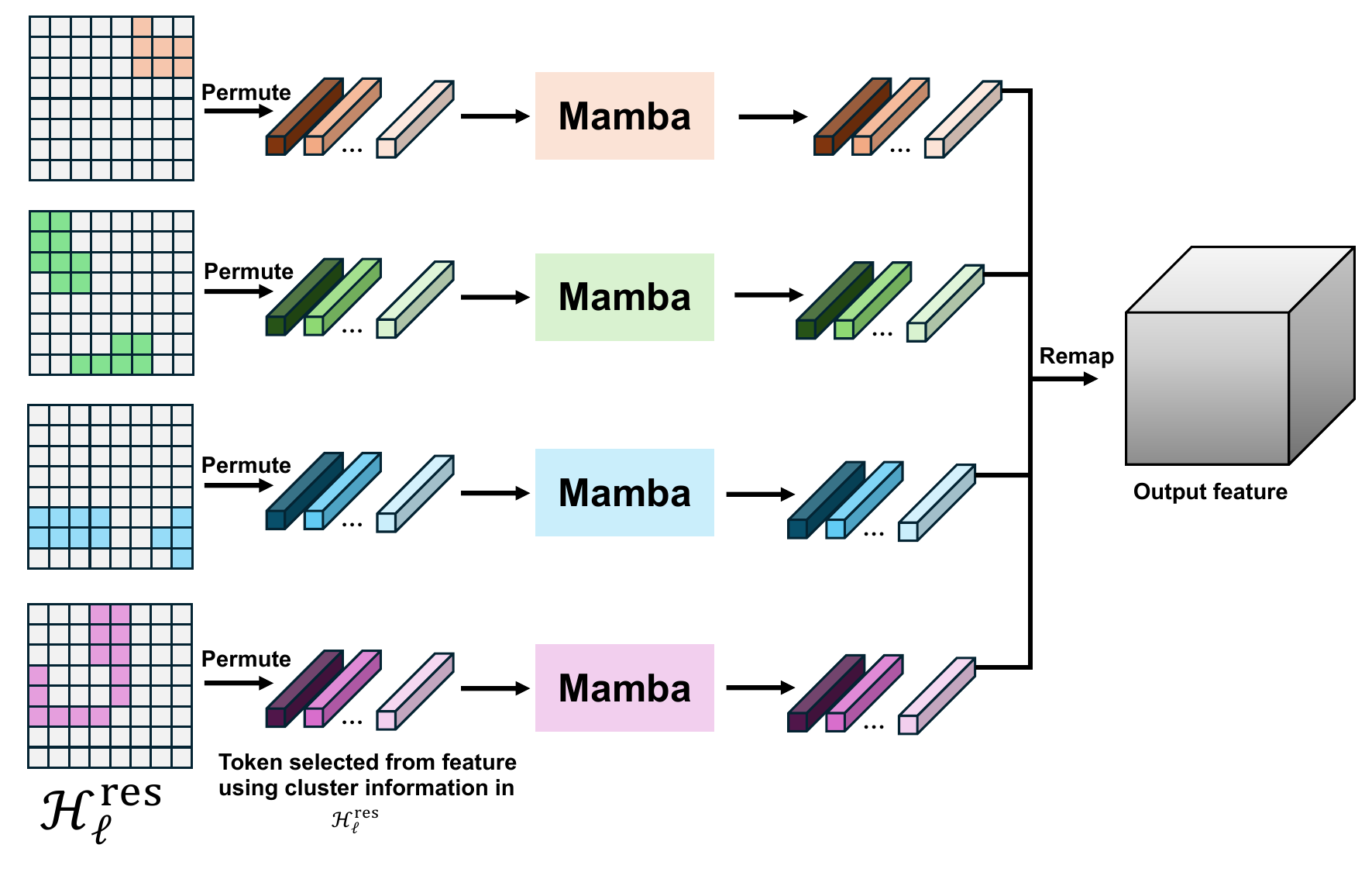}
    \caption{Illustration of cluster-wise Spatial Mamba block in layer function \(\mathcal{F}\). Clustering effect is found in \(\mathcal{H}^{\text{res}}\) and used for reducing the token and sequence length. Take the expansion rate \(n=2\) as an example.}
    \label{fig2}
\end{figure}

\subsection{Electromagnetic spectrum–aware residual stream}

\begin{figure}[htbp]
    \centering
    \includegraphics[width=0.49\textwidth]{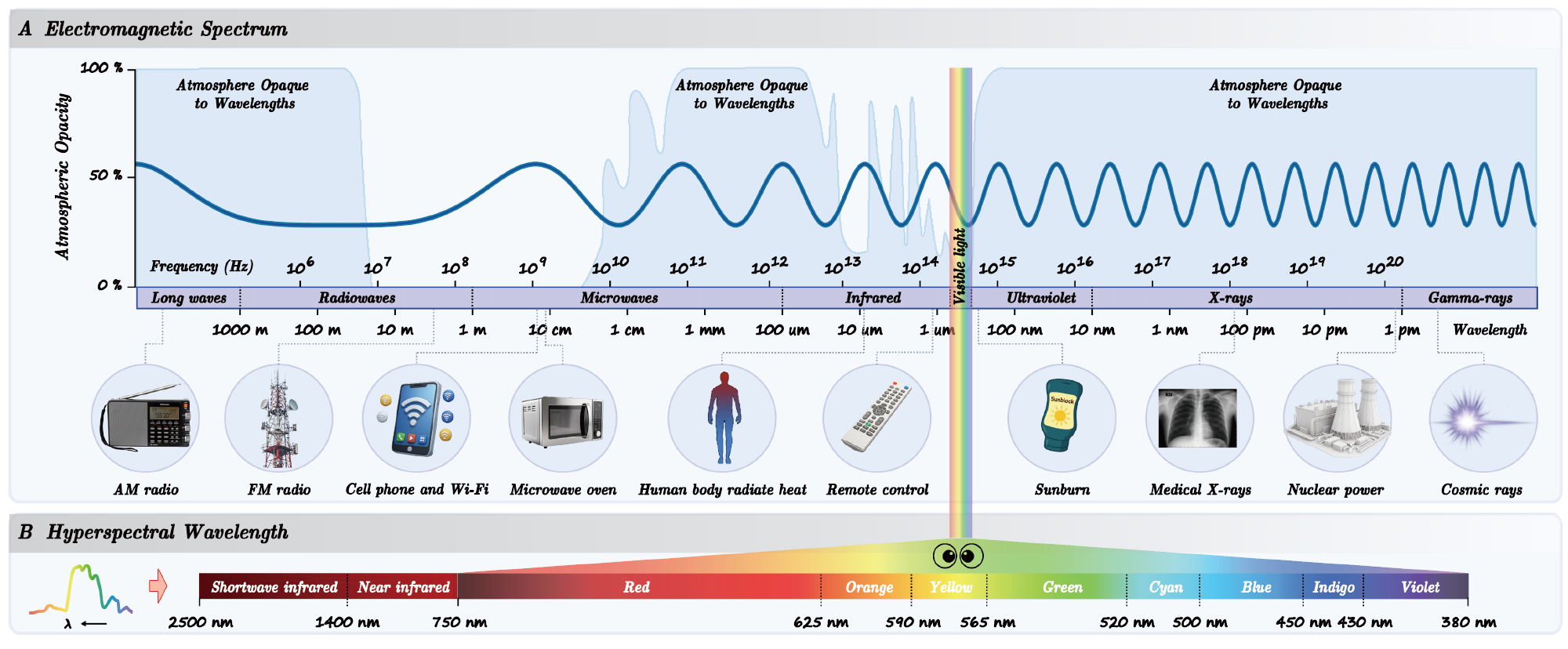}
    \caption{Overview of hyperspectral imaging. (A) Graphical illustration of the electromagnetic spectrum. (B) Expanded view of the typical wavelength regions captured in HSI: visible light (400-750 nm), near-infrared (NIR, 750-1400 nm), and shortwave infrared (SWIR, 1400-2500 nm). This figure comes from \cite{hong2025hyperspectral}.}
    \label{fig3}
\end{figure}

As shown in \autoref{fig3} \cite{hong2025hyperspectral}, HSI focuses on the optical window of the electromagnetic spectrum (see \autoref{fig3} A), typically covering wavelengths from 380 to 2500 nm. This window usually encompasses the visible light (400-700 nm), near-infrared (NIR), and shortwave infrared (SWIR) regions, as shown in \autoref{fig3} B. Each spectral range is sensitive to distinct material properties: visible bands capture surface color and pigment information, NIR reflects vegetation structure and health, and SWIR is strongly related to moisture content, soil composition, and burned materials. While the spectral responses vary across wavelengths, all bands observe the same underlying spatial structures, preserving consistent spatial patterns while encoding complementary physical information. This joint spatial coherence and spectral diversity underpin spatial–spectral analysis and spectral unmixing methods.

\begin{figure}[!h]
    \centering
    \includegraphics[width=0.49\textwidth]{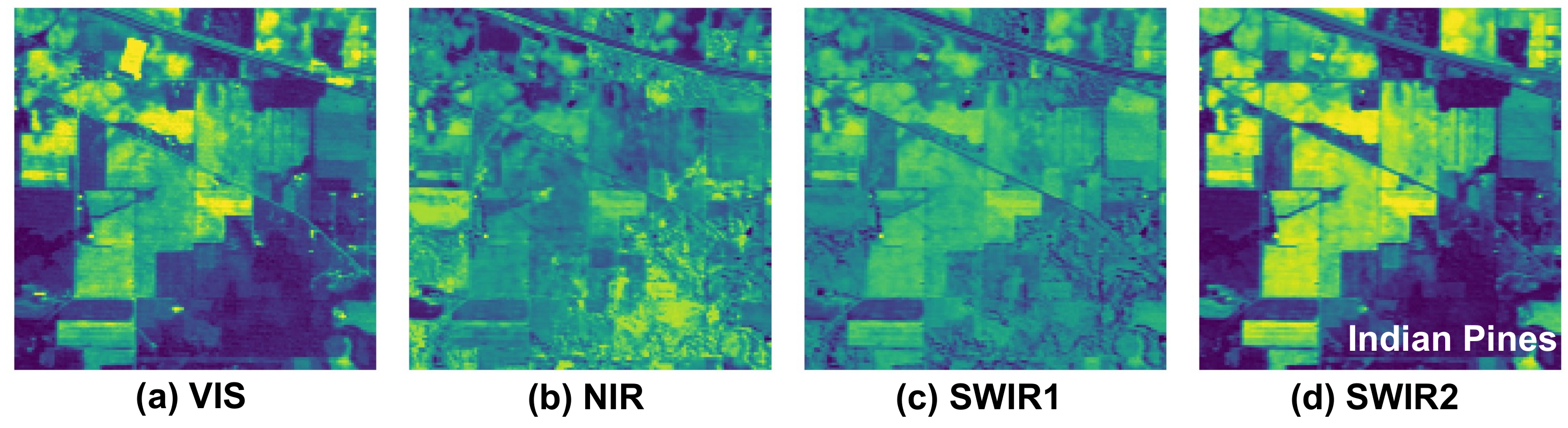}
    \caption{Illustration of the four electromagnetic spectrum–aware sub-cubes. (a) VIS, (b) NIR, (c) SWIR1, (d) SWIR2.}
    \label{fig4}
\end{figure}

Instead of replicating the feature like HC and \textit{m}HC did, forming expanded feature widths, to benefit the expanded connections, we fully consider the unique characteristic of the HSI cube from the electromagnetic spectrum perspective, by splitting the original HSI cube into non-overlapping sub-cubes to expand the width of the neural network's input feature, hence increasing the dense connectivity and multi-path structures, also maintain and enhances stability and scalability due to the manifold constraint of residual stream mixing matrix \(\mathcal{H}^{\text{res}}\). Two examples are shown in \autoref{fig4}. We can see that the spatial pattern and structure are well preserved, but the intensity reflected by the reflectance of different spectral ranges is different.

\subsection{Cluster-wise sequence scanning}
In the remote sensing field, the pixel-based and object-based image analysis are two main stream methods. One recent research study about the Sentinel-2 land use and cover utilizes the superpixel-based object-level approach to define the token in the Mamba model, which reduces model parameters and can also increase the classification accuracy \cite{dewis2025multitaskglocalobiamambasentinel2}. Superpixel is also used in the hyperspectral unmixing study \cite{shi2022deep}. \cite{ahmad2025graphmamba} also studied the permutation and connectivity of tokens in Mamba, but it is not a cluster-wise method, leading to limited consideration of spatial consistency.

In this paper, we found that in the residual stream mixing matrix \(\mathcal{H}^{\text{res}}\), there are clear and consistent clustering phenomena that will guide the token selection in the Mamba model, contributing to fewer but more related tokens. Additionally, this clustering phenomenon leads us to analyze the hidden connection of different clusters with the semantic label in the ground truth. 

The \autoref{fig2} gives the overview of the cluster-wise Mamba (CWM) scanning. Each colored cluster comes from one of the elements in \(\mathcal{H}^{\text{res}} \in \mathbb{R}^{L \times n \times n}, L = H \times W\). Supposing that here \(n=2\), so a total of four elements with size \(H \times W\). By selecting the Top-k tokens in each spatial matrix, \(n^2\) parallel spatial Mamba blocks are used to extract the spatial information for each element $(i,j)$ of $\mathcal{H}^{\text{res}}$, which can be expressed as the following:
\begin{align}
    \begin{aligned}
        \mathcal{T}^{i,j} &= \operatorname{Top\text{-}k}\!\left(\boldsymbol{R}, \mathcal{H}^{\text{res}}_{:,i,j}\right), \\
        \hat{\mathcal{T}}^{i,j} &= \operatorname{CWM}_{i,j}\!\left(\mathcal{T}^{i,j}\right),
        \quad i,j \in \{1,\dots,n\},\\
    \end{aligned}
\end{align}
where $\operatorname{CWM}_{i,j}$ denotes a cluster-wise Mamba applied in parallel across all $(i,j)$ components. \(\boldsymbol{R}\) is the feature map at layer \(l\). After the parallel CWM blocks are finished, all the tokens are remapped to the original location in feature \(\boldsymbol{R}\), as follows:
\begin{align}
    \begin{aligned}
        \hat{\boldsymbol{R}_{l}} &= \boldsymbol{R}_{l} + \textbf{Map}((\operatorname{sort}^{-1}(\hat{\mathcal{T}}^{i,j}))
    \end{aligned}
\end{align}
\(\operatorname{sort}^{-1}\) represents to recover to the original order. \textbf{Map} means put the processed feature at the original spatial location.

\subsection{Spectral-Spatial Mamba Block}

Cluster-wise Mamba is one part of the Spectral-Spatial Mamba block for spatial information representation. By splitting and grouping the input feature \(\boldsymbol{R}\) along the channel dimension, forming the input data of the spectral Mamba, each group is viewed as a token to be processed, as follows:
\begin{align}
    \begin{aligned}
        \boldsymbol{R}_l = \boldsymbol{R}_l + \text{Reshape} (\textbf{Mamba}(\text{ChannelSplit}(\boldsymbol{R}_l))
    \end{aligned}
\end{align}

The algorithm \autoref{algo1} shows the pipeline of our proposed model, where SSM is the spectral-spatial Mamaba block and FFN is the feed forward layer. \(\mathcal{H}^{res}_{l} \in \mathbb{R}^{L \times n \times n}, \mathcal{H}^{\text{post}}_{l} \in \mathbb{R}^{L \times n}, \mathcal{H}^{\text{pre}}_{l} \in \mathbb{R}^{L \times n}\). Overall, in the feature layer \(\mathcal{F}\), shown in \autoref{fig1} (A), has two types module, one is the proposed spectral-spatial Mamba block with cluster-wise spatial Mamba and spectral Mamba, and another is the FFN. In each type, the manifold constrained mapping is used to replace the residual connection to realize the identity mapping. Spatial positional encoding is injected into the full-spectrum stream, which serves as a spatial anchor. Other spectral streams receive spatial context implicitly through hyper-connection interactions.
\begin{algorithm}[htbp]
\caption{Our proposed model}
    \begin{algorithmic}[1] \label{algo1}
        \REQUIRE Hyperspectral image cube \(\mathbf{H} \in \mathbb{R}^{H \times W \times C}\), training samples mask set \(\mathcal{D} \in \mathbb{R}^{H \times W}\), four defined electromagnetic spectrum–aware residual stream with additional full bands \(\mathbf{E} \in \{\text{FULL}, \text{VIS}, \text{NIR}, \text{SWIR1}, \text{SWIR2}\}\) in \autoref{method}. Hidden dimension \(D\). Expansion rate \(n = 1+4=5\). Total \(L\) layers in \textit{m}HC blocks set \(\mathcal{L}\)
         \STATE initialize a residual stream list \(\mathcal{RS}\)
         \FOR{\(e\) \text{in} \(\mathbf{E}\)}
         \STATE get the corresponding cube \(\mathbf{H}_e \in \mathbf{R}^{H \times W \times C_e}\) for physical band \(e\)
         \STATE \(\mathbf{F}_{C_e}\) = \textbf{Embedding}(\(\mathbf{H}_e\))
         \IF {\(e = \text{FULL}\)}
         \STATE \(\mathbf{F}_{C_e} = \mathbf{F}_{C_e} + \text{Spatial Position Encoding}\) 
         \ENDIF
        \STATE append \(\mathbf{F}_{C_e}\) to \(\mathcal{RS}\)
        \ENDFOR
        \STATE get the residual stream \(\boldsymbol{R} \in \mathbb{R}^{L \times n \times D}\) by stacking \(\mathcal{RS}\) together 
        \FOR{\textit{m}HC layer \(l\) in block sets \(\mathcal{L}\)}
        \STATE \(\tilde{\boldsymbol{R}_{l}}\) = \text{RMSNorm} (\(\boldsymbol{R}_{l}\))
        \STATE \(\tilde{\mathcal{H}}_l^{\text{pre}} = \alpha_{l}^{\text{pre}} \cdot \text{tanh}(\theta_{l}^{\text{pre}} \tilde{\boldsymbol{R}}_{l}^T) + \mathbf{b}_{l}^{\text{pre}}\) 
        \STATE \(\tilde{\mathcal{H}}_{l}^{\text{post}} = \alpha_{l}^{\text{post}} \cdot \text{tanh} (\theta_{l}^{\text{post}}\tilde{\boldsymbol{R}}_{l}^T) + \mathbf{b}_{l}^{\text{post}}\)
        \STATE \(\tilde{\mathcal{H}}_{l}^{\text{res}} = \alpha_{l}^{\text{res}} \cdot \text{tanh} (\theta_{l}^{\text{res}}\tilde{\boldsymbol{R}}_{l}^T) + \mathbf{b}_{l}^{\text{res}}\)
        \STATE \(\mathcal{H}^{\text{pre}}_{l} = \sigma (\tilde{\mathcal{H}}^{\text{pre}}_{l})\)
        \STATE \(\mathcal{H}^{\text{post}}_{l} = 2\sigma (\tilde{\mathcal{H}}^{\text{post}}_{l})\)
        \STATE \(\mathcal{H}^{\text{res}}_{l} = \text{Sinkhorn-Knopp} (\tilde{\mathcal{H}}^{\text{res}}_{l})\)
        \STATE \(\hat{\boldsymbol{R}_{l}} = \mathcal{H}_{l}^{\text{res}} \boldsymbol{R}_{l} + \mathcal{H}_{l}^{\text{post}} (\text{SSM}_{l}(\mathcal{H}^{\text{pre}}_{l} \boldsymbol{R}_{l}))\)
        \STATE \(\bar{\boldsymbol{R}_{l}} = \text{RMSNorm}(\hat{\boldsymbol{R}_{l}})\)
        \STATE \(\bar{\mathcal{H}}_l^{\text{pre}} = \alpha_{l}^{\text{pre}} \cdot \text{tanh}(\theta_{l}^{\text{pre}} \bar{\boldsymbol{R}}_{l}^T) + \mathbf{b}_{l}^{\text{pre}}\) 
        \STATE \(\bar{\mathcal{H}}_{l}^{\text{post}} = \alpha_{l}^{\text{post}} \cdot \text{tanh} (\theta_{l}^{\text{post}}\bar{\boldsymbol{R}}_{l}^T) + \mathbf{b}_{l}^{\text{post}}\)
        \STATE \(\bar{\mathcal{H}}_{l}^{\text{res}} = \alpha_{l}^{\text{res}} \cdot \text{tanh} (\theta_{l}^{\text{res}}\bar{\boldsymbol{R}}_{l}^T) + \mathbf{b}_{l}^{\text{res}}\)
        \STATE \(\mathcal{H}^{\text{pre}}_{l} = \sigma (\bar{\mathcal{H}}^{\text{pre}}_{l})\)
        \STATE \(\mathcal{H}^{\text{post}}_{l} = 2\sigma (\bar{\mathcal{H}}^{\text{post}}_{l})\)
        \STATE \(\mathcal{H}^{\text{res}}_{l} = \text{Sinkhorn-Knopp} (\bar{\mathcal{H}}^{\text{res}}_{l})\) 
        \STATE \({\boldsymbol{R}_{l+1}} = \mathcal{H}_{l}^{\text{res}} \hat{\boldsymbol{R}}_{l} + \mathcal{H}_{l}^{\text{post}} (\text{FFN}_{l}(\mathcal{H}^{\text{pre}}_{l} \hat{\boldsymbol{R}}_{l}))\)
        \ENDFOR
        \STATE final feature \(h = \text{Mean}(\boldsymbol{R}_{L}, \text{dim}=2) \in \mathbb{R}^{L \times D}\)
        \STATE run a classification head on \(h\) to get the prediction logits
        \STATE calculate the cross-entropy using training samples
        \STATE update model parameters and repeat
    \end{algorithmic}
\end{algorithm}

\section{Experiments} \label{exp}

\begin{table*}[!h]
\caption{Quantitative performance of different classification methods in terms of OA, AA, $k$, as well as the accuracies for each class on the Indian Pines dataset with 10 \% training samples. The best results are in bold and colored shadow.}
\resizebox{\textwidth}{!}{
\begin{tabular}{ccc|cc|c|ccc|cc|c}
\hline
\multicolumn{1}{c|}{\multirow{2}{*}{Class No.}} & \multicolumn{1}{c|}{\multirow{2}{*}{Train Number}} & \multirow{2}{*}{Test Number} & \multicolumn{2}{c|}{CNN-based} & GAN-based & \multicolumn{3}{c|}{Transformer-based} & \multicolumn{2}{c|}{Mamba-based} &  \textit{m}HC-based                      \\ \cline{4-12} 
\multicolumn{1}{c|}{}                           & \multicolumn{1}{c|}{}                              &                              & SSRN      & SS-ConvNeXt        & MTGAN     & SSFTT  & SSTN         & GSC-ViT        & MambaHSI      & 3DSS-Mamba & \textbf{Ours}  \\ \hline
\multicolumn{1}{c|}{1}                          & \multicolumn{1}{c|}{5}                             & 37                           & 90.98     & 84.21              & 84.63     & 94.18  & 94.44        & 94.59          & 94.74         & 95.12      & \cellcolor[RGB]{251, 228, 213}\textbf{100.00}           \\
\multicolumn{1}{c|}{2}                          & \multicolumn{1}{c|}{143}                           & 1142                         & 97.54     & \cellcolor[RGB]{251, 228, 213}\textbf{99.53}     & 97.20      & 94.9   & 97.99        & 95.19          & 98.08                 & 93.22      & 99.13          \\
\multicolumn{1}{c|}{3}                          & \multicolumn{1}{c|}{83}                            & 664                          & 96.88     & 96.62              & 95.9      & 92.16  & 95.48        & 96.54          & 96.36               & 93.44      & \cellcolor[RGB]{251, 228, 213}\textbf{98.24}  \\
\multicolumn{1}{c|}{4}                          & \multicolumn{1}{c|}{24}                            & 189                          & 97.7      & 95.24              & 96.85     & 94.8   & 93.12        & \cellcolor[RGB]{251, 228, 213}\textbf{99.47} & 90.05                & 96.71      & \cellcolor[RGB]{251, 228, 213}\textbf{100.00}   \\
\multicolumn{1}{c|}{5}                          & \multicolumn{1}{c|}{48}                            & 387                          & 95.13     & 97.91              & 95.61     & 95.33  & \cellcolor[RGB]{251, 228, 213}\textbf{99.22}        & 98.45          & 97.67       & 96.71      & 95.82          \\
\multicolumn{1}{c|}{6}                          & \multicolumn{1}{c|}{73}                            & 584                          & 99.25     & 99.85              & 98.48     & 96.66  & 99.66        & 98.29          & 99.49       & 95.40      & \cellcolor[RGB]{251, 228, 213}\textbf{100.00}          \\
\multicolumn{1}{c|}{7}                          & \multicolumn{1}{c|}{3}                             & 22                           & 76.4      & \cellcolor[RGB]{251, 228, 213}\textbf{100.00}                & 12.00        & 84.92  & 68.18        & 77.27          & \cellcolor[RGB]{251, 228, 213}\textbf{100.00}    & 98.63      & 91.67   \\
\multicolumn{1}{c|}{8}                          & \multicolumn{1}{c|}{48}                            & 382                          & 99.53     & 99.76              & 99.95     & 99.62  & \cellcolor[RGB]{251, 228, 213}\textbf{100.00}          & \cellcolor[RGB]{251, 228, 213}\textbf{100.00}            & \cellcolor[RGB]{251, 228, 213}\textbf{100.00}   & 96.97      & \cellcolor[RGB]{251, 228, 213}\textbf{100.00}   \\
\multicolumn{1}{c|}{9}                          & \multicolumn{1}{c|}{2}                             & 16                           & 55.29     & 92.86              & 61.76     & 77.83  & 78.57        & 93.75          & \cellcolor[RGB]{251, 228, 213}\textbf{100.00}           & 72.22      & \cellcolor[RGB]{251, 228, 213}\textbf{100.00}   \\
\multicolumn{1}{c|}{10}                         & \multicolumn{1}{c|}{97}                            & 778                          & 96.19     & 97.81              & 95.43     & 92.21  & \cellcolor[RGB]{251, 228, 213}\textbf{98.97}        & 97.17          & 97.04   & 92.80      & 97.34 \\
\multicolumn{1}{c|}{11}                         & \multicolumn{1}{c|}{245}                           & 1965                         & 98.29     & 88.78              & 98.44     & 97.52  & 97.91        & 99.75          &\cellcolor[RGB]{251, 228, 213}\textbf{ 99.29 }               & 96.92      & 99.13 \\
\multicolumn{1}{c|}{12}                         & \multicolumn{1}{c|}{59}                            & 475                          & 97.97     & 97.92              & 95.10     & 90.48  & \cellcolor[RGB]{251, 228, 213}\textbf{100.00} & 95.16          & 99.16              & 90.64      & 96.79          \\
\multicolumn{1}{c|}{13}                         & \multicolumn{1}{c|}{20}                            & 164                          & 99.68     & \cellcolor[RGB]{251, 228, 213}\textbf{100.00}       & 98.76     & 96.59  & 99.39        & \cellcolor[RGB]{251, 228, 213}\textbf{100.00}   & 99.39                & 95.14      & \cellcolor[RGB]{251, 228, 213}\textbf{100.00}   \\
\multicolumn{1}{c|}{14}                         & \multicolumn{1}{c|}{126}                           & 1012                         & 99.57     & \cellcolor[RGB]{251, 228, 213}\textbf{99.91}              & 99.12     & 98.67  & 99.8         & 99.7           & 99.11              & 99.65      & \cellcolor[RGB]{251, 228, 213}\textbf{99.91}   \\
\multicolumn{1}{c|}{15}                         & \multicolumn{1}{c|}{39}                            & 308                          & 98.01     & 99.71              & 98.10     & 83.48  & 99.35        & 96.44          & \cellcolor[RGB]{251, 228, 213}\textbf{100.00}     & 95.68      & \cellcolor[RGB]{251, 228, 213}\textbf{100.00}   \\
\multicolumn{1}{c|}{16}                         & \multicolumn{1}{c|}{9}                             & 75                           & 97.38     & 97.53              & 92.50     & 89.93  & 89.33        & 97.3           & 85.33             & 85.71      & \cellcolor[RGB]{251, 228, 213}\textbf{98.79} \\ \hline
\multicolumn{3}{c|}{OA(\%)}                                                                                                         & 97.81     & 96.29              & 97.20     & 95.12  & 98.23        & 97.94          & 98.28              & 95.53      & \cellcolor[RGB]{251, 228, 213}\textbf{98.85} \\
\multicolumn{3}{c|}{AA(\%)}                                                                                                         & 93.49     & 96.73              & 88.80     & 92.46  & 94.49        & 96.19          & 97.23             & 94.90       & \cellcolor[RGB]{251, 228, 213}\textbf{98.55} \\
\multicolumn{3}{c|}{Kappa(\%)}                                                                                                      & 97.74     & 95.96              & 97.08     & 94.76  & 97.98        & 97.65          & \cellcolor[RGB]{251, 228, 213}\textbf{98.21}              & 93.42      & 97.44 \\ \hline
\end{tabular}
\label{table:IP}
}
\end{table*}

\subsection{Datasets Description}
\subsubsection{\textbf{Indian Pines Data}} This dataset was collected by the AVIRIS sensor over Northwestern Indiana, USA. This data consists of 145 × 145 pixels at a ground sampling distance (GSD) of 20 m and 220 spectral bands covering the wavelength range of 400–2500 nm with a 10-m spectral resolution. In the experiment, 24 water-absorption bands and noise bands were removed, and 200 bands were selected. There are 16 mainly investigated categories in this studied scene. Since the wavelength range is from 400–2500 nm, four groups are split, as shown in \autoref{fig4}.

% \subsubsection{\textbf{Liao Ning-01}} This real data, LN01, comes from the ZY-1(02D) satellite, China \cite{jia2025spectral}. It carries both hyperspectral and multispectral payloads, which can capture hyperspectral and MSI at the same scene. The hyperspectral payload mainly comprises a geostationary hyperspectral imager, which can capture high-resolution and continuous-spectrum satellite images. The hyperspectral image covers a spectral range from 0.4 to 2.5 \(\mu m\), consisting of 166 bands, \(300 \times 300\) pixels, including 76 bands in the visible and near-infrared and 90 bands in the short-wave infrared. This area is formed by the confluence of seawater and inland waters, featuring a variety of land cover types, including reservoirs, seawater, sandy soil, arable land, bare soil, highways, railways, barren grass, mountain vegetation, and a broken bridge. Since the wavelength range is from 400 to 2500 nm, four groups are split, as shown in \autoref{fig4}. Additionally, the size of ground truth is \(900 \times 900\), so majority downsampling is used.

\subsection{Classification Results}

\autoref{table:IP} shows the numerical results achieved by different methods on the Indian Pines dataset. Our approach outperforms the other methods on all metrics. In particular, our approach ES-\textit{m}HC, achieves much better results on AA, indicating that the proposed approach outperforms the other approaches in terms of preserving and classifying the small classes.

\begin{figure}[!h]
    \centering
    \includegraphics[width=0.49\textwidth]{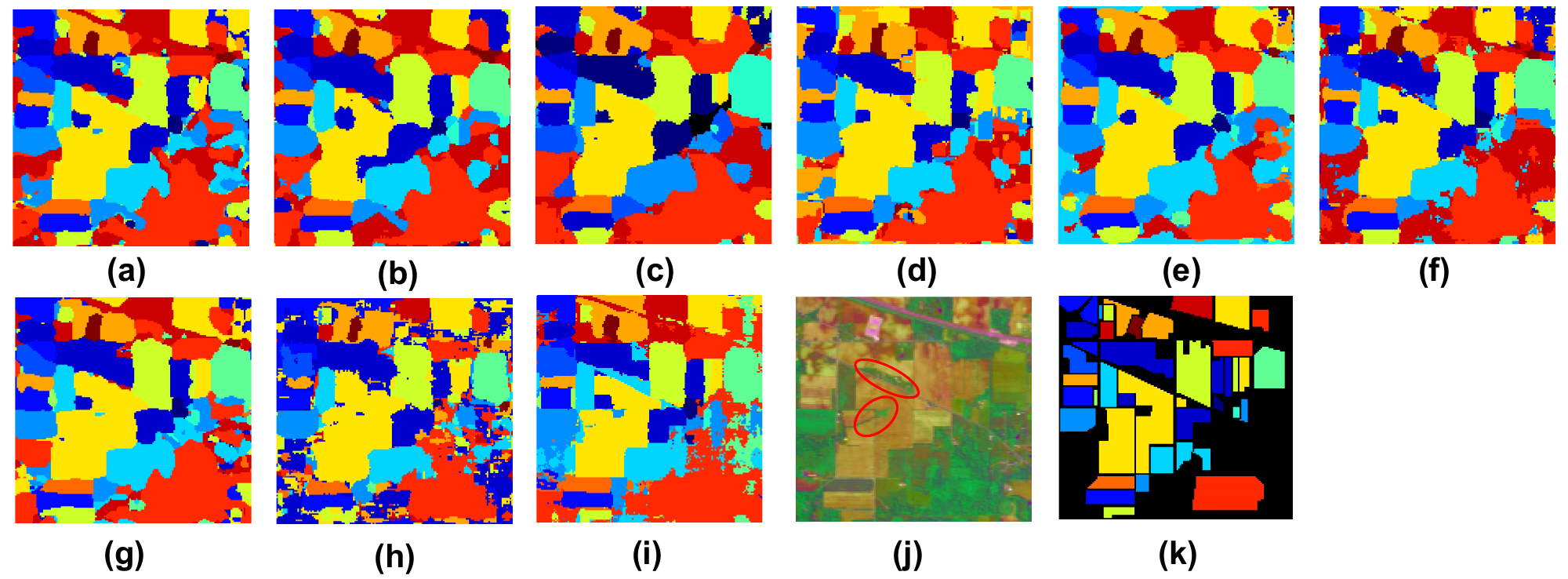}\\
    \includegraphics[width=0.49\textwidth]{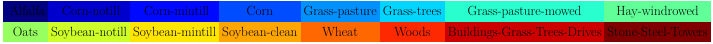}
    \caption{The Indian Pines classification map generated by different methods. (a) SSRN (b) SS-ConvNeXt (c) MTGAN (d) SSFTT (e) SSTN (f) GSC-ViT (g) MammbaHSI (h) 3DSS-Mamba (i) ES-\textit{m}HC (j) False Color Image (k) Ground Truth. Some red circles are shown on the  RGB image to illustrate the boundary preservation of our proposed model.}
    \label{maps}
\end{figure}

Additionally, as shown in \autoref{maps}, the proposed approach achieves a map that is not only the most consistent with the classification map, but also better at delineating the boundaries and small classes, as shown in the two red circles ROI areas.
\subsection{Impact on the expansion rate}

\begin{figure}[!h]
    \centering
    \includegraphics[width=0.49\textwidth]{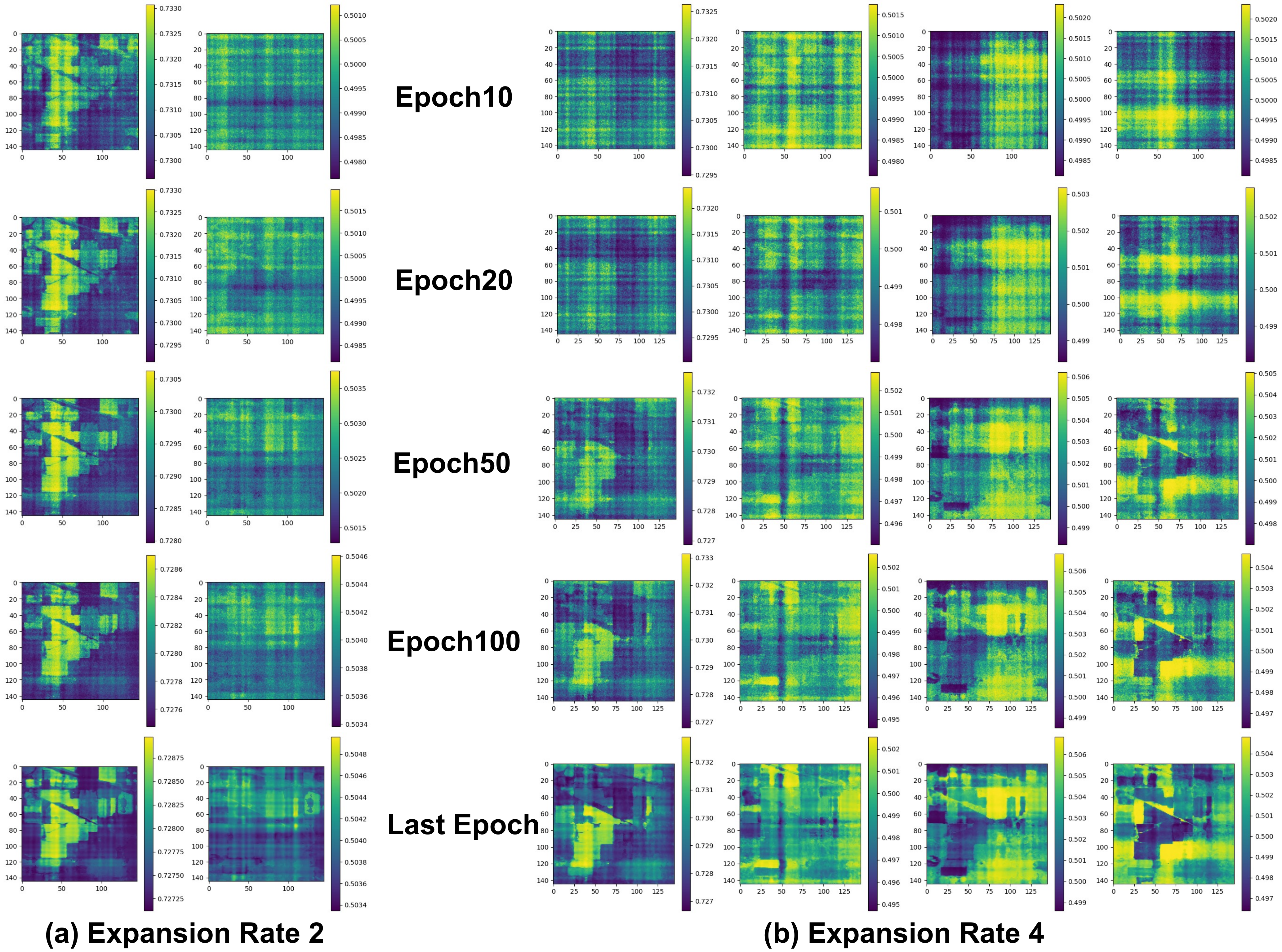}
    \caption{Visualization of the \(\mathcal{H}^{\text{pre}}\) at different epoch and expansion rate.}
    \label{H_pre}
\end{figure}

\begin{figure*}[!h]
    \centering
    \includegraphics[width=\textwidth]{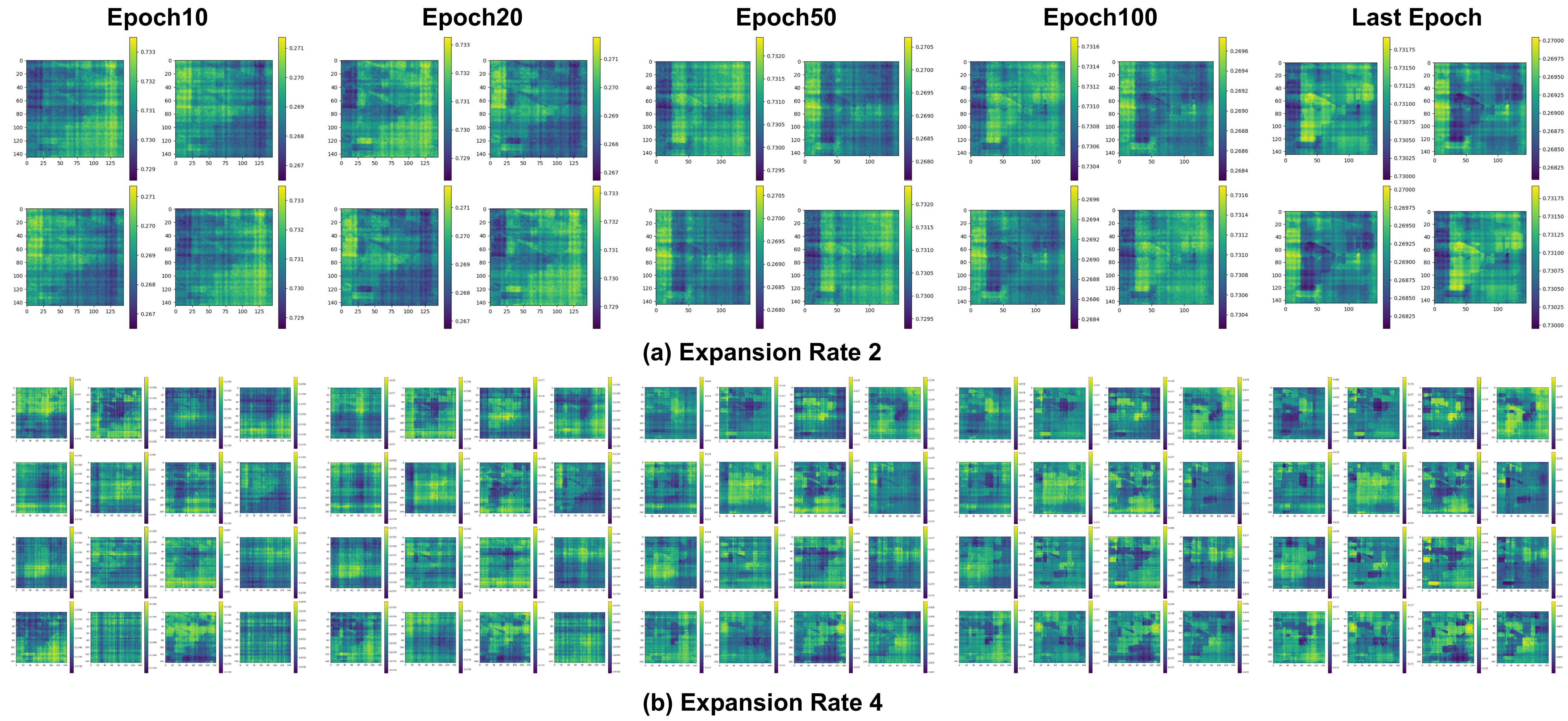}
    \caption{Visualization of the \(\mathcal{H}^{\text{res}}\) at different epoch and expansion rate.}
    \label{H_res_compare}
\end{figure*}

\begin{figure}[!h]
    \centering
    \includegraphics[width=0.49\textwidth]{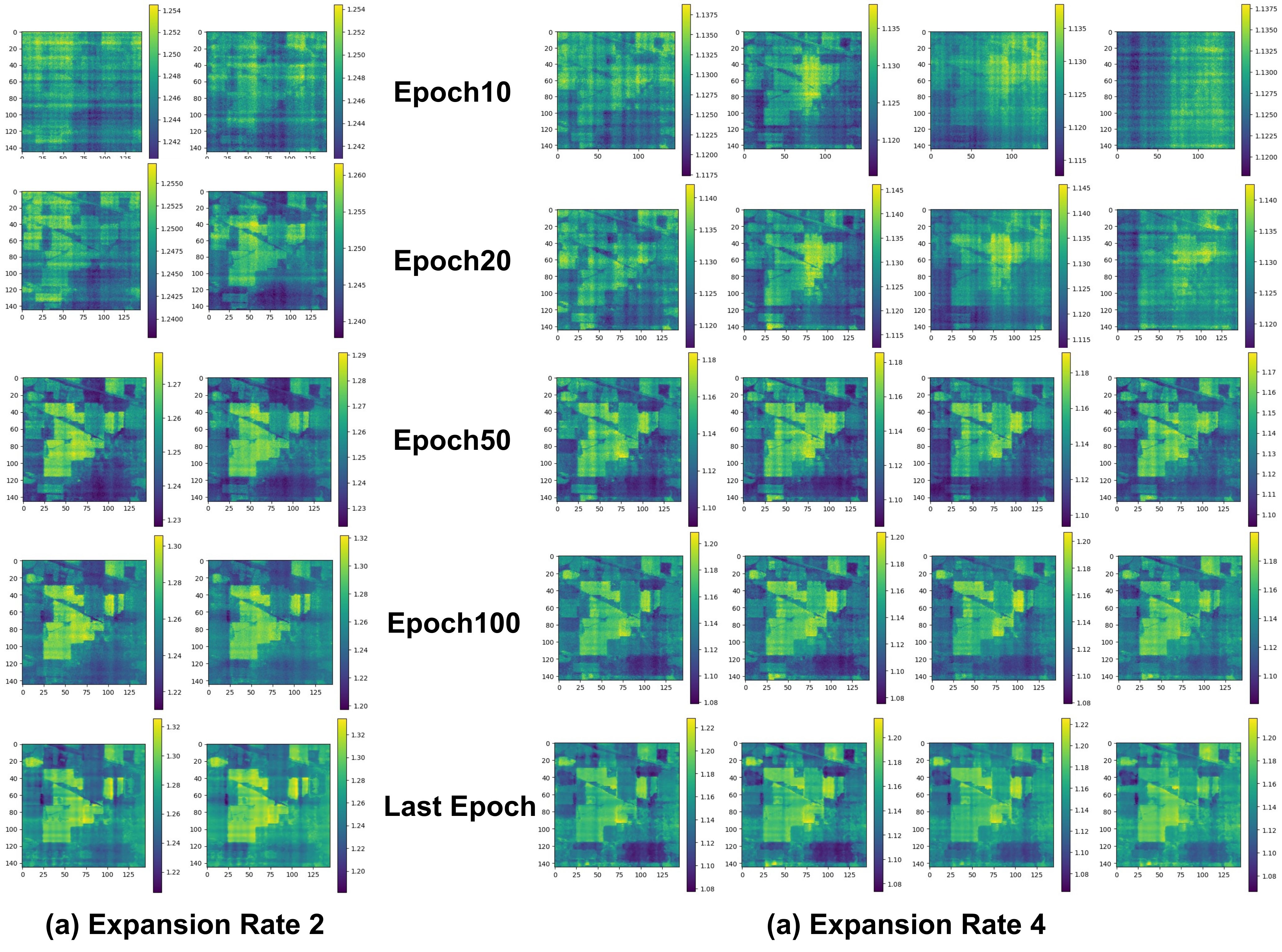}
    \caption{Visualization of the \(\mathcal{H}^{\text{post}}\) at different epoch and expansion rate.}
    \label{H_post}
\end{figure}

\begin{table}[!h]
\caption{Impact on the expansion rate \(n\).}
\resizebox{0.49\textwidth}{!}{
\begin{tabular}{c|cc}
\hline
Class No. & Expansion rate \(n\)=2 & Expansion rate \(n\)=4 \\ \hline
1         & 95.12              & 97.56              \\
2         & 98.74              & 98.19              \\
3         & 98.78              & 99.05              \\
4         & 99.05              & 96.69              \\
5         & 95.59              & 95.82              \\
6         & 99.84              & 99.84              \\
7         & 83.33              & 87.50              \\
8         & 100.00             & 100.00             \\
9         & 100.00             & 100.00             \\
10        & 97.69              & 96.42              \\
11        & 99.81              & 99.31              \\
12        & 96.97              & 95.27              \\
13        & 100.00             & 100.00             \\
14        & 100.00             & 100.00             \\
15        & 100.00             & 100.00             \\
16        & 97.59              & 98.80              \\ \hline
OA        & 98.95              & 98.54              \\
AA        & 97.65              & 97.77              \\
Kappa     & 95.98              & 95.80              \\ \hline
\end{tabular}
}
\label{expansion_rate}
\end{table}
Furthermore, we explore the impact on the expansion rate, and visualize the three key matrices, \(H_{\text{res}}\), \(\mathcal{H}^{\text{post}}\), \(\mathcal{H}^{\text{pre}}\). Note that this experiment is under the setting of duplicating the input feature, i.e., deplicating the original HSI cube for \(n\) times as \textit{m}HC did, instead of split into more physical meaningful spctrum bands, because the visible light (VIS, 400-700 nm), near-infrared (NIR, 700-1000 nm), and shortwave infrared 1 (SWIR1, 1000-1800 nm) and shortwave infrared 2 (SWIR2, 1800-2500 nm), are all well-predefined. The numerical results are shown in \autoref{expansion_rate}, which demonstrate that the comparable classification performance when only deplicate the input HSI cube, but using electromagnetic spectrum–aware resifual stream design method, the results are better than original deplication for network's input width expansion. From multi-view learning perspective, our proposed ES-\textit{m}HC has more clear physical meaning which can help us to understand the model inference.

To understand the model behavior, we visualize the three key matrices, \(\mathcal{H}^{\text{pre}}\), \(\mathcal{H}^{\text{post}}\), \(\mathcal{H}^{\text{res}}\) for expansion rate 2 and 4. The visual results are shown as follows in \autoref{H_pre}, \autoref{H_post}, and \autoref{H_res_compare}. As illustrated in \textit{m}HC paper, the Figure 1 \cite{xie2026mhc}, \(\mathcal{H}^{\text{pre}}\) serves as the role of the feature compression, by compressing the expanded \(n\) features into one representative feature. While, \(\mathcal{H}^{\text{post}}\) serves as feature reconstruction matrix to map the compressed feature to the original size. \(\mathcal{H}^{\text{res}}\) is the learnable mapping that mixes features within the \(n\) residual streams. More importantly, these three matrices are learned from data, which means they are feature-dependent parameters, see \autoref{algo1} and \textit{m}HC paper \cite{xie2026mhc}.

As shown in \autoref{H_pre}, when the expansion rate is 2, the expanded stream (i.e., the second column) is harder to learn the spatial pattern than the main stream (i.e., the first column). While, when expansion rate is 4, the spatial pattern within each stream (i.e., each column) appears to become clear during the training process. This could be explained by the more residual stream making the compressed feature more representative, leading to quick convergence of the spatial pattern.

\(\mathcal{H}^{\text{post}}\) reconstruct the compressed feature. As we can see from \autoref{H_post}, at epoch 50, all the learnable mapping matrices can well preserve the spatial pattern. \autoref{H_res_compare} reflects the information transition among each residual stream. When the expansion rate is 2, meaning a narrower network input, limited representation, leading to unclear and not well-preserved spatial patterns. Compared with \(n=2\), bigger expansion rate \(n=4\) show clear spatial pattern. Although some learnable mapping matrix is still blurry, the emergence of spatial patterns is quicker than the small expansion rate.

\begin{figure*}[!h]
    \centering
    \includegraphics[width=\textwidth]{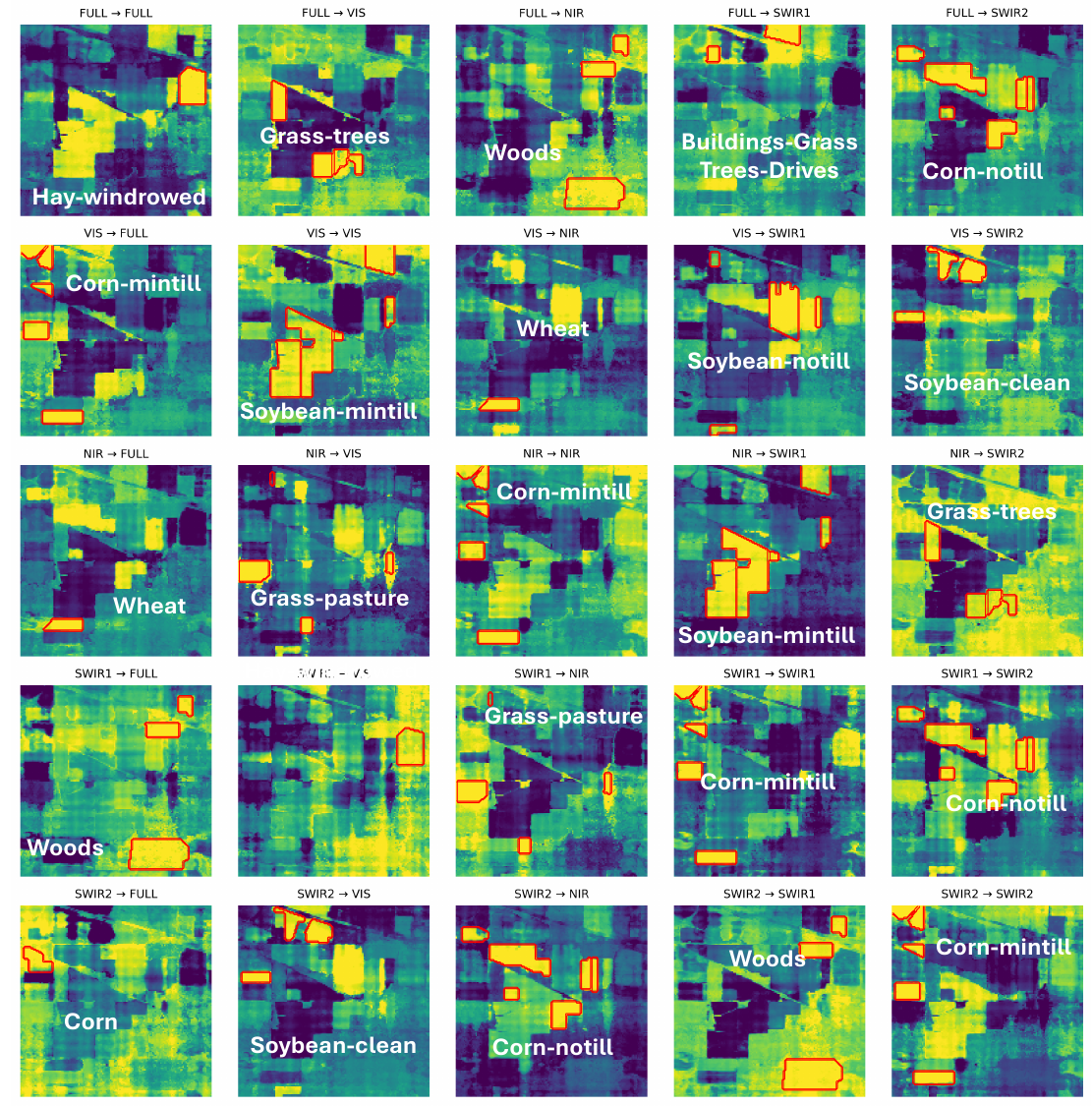}
    \caption{Visualization of the interpretable \(\mathcal{H}^{\text{res}}\) with overlaid class boundaries. The boundary is selected based on the highest mean value for each class boundary. White text shows the name of the category.}
    \label{H_res}
\end{figure*}
\autoref{H_res} shows the correspondence between high-value regions in \(\mathcal{H}^{\text{res}}\) and category boundary. These results are based on the ES-\textit{m}HC. As we can see from this figure, each learnable mapping matrix, a total of 25 matrices, has its own unique category tendency. Additionally, the bi-directional stream flow is asymmetric, also shown in the \autoref{fig1} (B), the joint distribution of the off-diagonal element. For example, "FULL \(\rightarrow\) VIS" flow shows the high-value area in the Grass-trees (class 6), while "VIS \(\rightarrow\) FULL" show high value in the Corn-mintill area. "Corn-notill, Corn-mintill, Corn" are more related to the SWIR bands, as they show higher values in "FULL \(\rightarrow\) SWIR2", "SWIR1 \(\rightarrow\) SWIR1", "SWIR1 \(\rightarrow\) SWIR2", "SWIR2 \(\rightarrow\) FULL", "SWIR2 \(\rightarrow\) NIR", and "SWIR2 \(\rightarrow\) SWIR2". The "notill" and "mintill" mean crop residue covers the surface, or the mixture of soil and crop residue, which could lead to lower reflectance of "Corn-notill" and "Corn-mintill" at SWIR.

\section{Discussion and Conclusion} \label{discussion}
In this paper, to our knowledge, we are the first one to propose an electromagnetic spectrum–aware residual stream splitting method to replace the \textit{m}HC residual design, and apply it to the hyperspectral image classification task. This physically meaningful way can be explained and can increase classification performance. In order to explain the mechanism of \textit{m}HC, making ES-mHC transforms HSIC into a structurally transparent, partially white-box learning process, we found that there are clear clustering effects in \(\mathcal{H}^{res}\) which motivate us to use the cluster-wise spatial Mamba block. Additionally, we visualized the \(\mathcal{H}^{res}\) and analyzed the high-value in residual stream information transition with the corresponding category.

The physically meaningful stream interactions can be viewed as different sensors, and we confirm that the design of \textit{m}HC will benefit the feature fusion in the feature. One of the potential reasons is that the non-negative feature of the learnable mapping \(\mathcal{H}^{\text{res}}\) has a positive effect on information fusion. In the feature, more explainable methods need to be proposed to analyze the three matrices in \textit{m}HC, because these three components are learned from the data, instead of the global parameter in the model, like the CNN kernel.

\bibliographystyle{IEEEtran}
\bibliography{IEEEabrv,references}

% Generated by IEEEtran.bst, version: 1.14 (2015/08/26)
\begin{thebibliography}{10}
\providecommand{\url}[1]{#1}
\csname url@samestyle\endcsname
\providecommand{\newblock}{\relax}
\providecommand{\bibinfo}[2]{#2}
\providecommand{\BIBentrySTDinterwordspacing}{\spaceskip=0pt\relax}
\providecommand{\BIBentryALTinterwordstretchfactor}{4}
\providecommand{\BIBentryALTinterwordspacing}{\spaceskip=\fontdimen2\font plus
\BIBentryALTinterwordstretchfactor\fontdimen3\font minus \fontdimen4\font\relax}
\providecommand{\BIBforeignlanguage}[2]{{%
\expandafter\ifx\csname l@#1\endcsname\relax
\typeout{** WARNING: IEEEtran.bst: No hyphenation pattern has been}%
\typeout{** loaded for the language `#1'. Using the pattern for}%
\typeout{** the default language instead.}%
\else
\language=\csname l@#1\endcsname
\fi
#2}}
\providecommand{\BIBdecl}{\relax}
\BIBdecl

\bibitem{rodarmel2002principal}
C.~Rodarmel and J.~Shan, ``Principal component analysis for hyperspectral image classification,'' \emph{Surveying and Land Information Science}, vol.~62, no.~2, pp. 115--122, 2002.

\bibitem{stone2002independent}
J.~V. Stone, ``Independent component analysis: an introduction,'' \emph{Trends in cognitive sciences}, vol.~6, no.~2, pp. 59--64, 2002.

\bibitem{SSRN}
Z.~Zhong, J.~Li, Z.~Luo, and M.~Chapman, ``Spectral–spatial residual network for hyperspectral image classification: A 3-d deep learning framework,'' \emph{IEEE Transactions on Geoscience and Remote Sensing}, vol.~56, no.~2, pp. 847--858, 2018.

\bibitem{mei2024novel}
S.~Mei, Z.~Han, M.~Ma, F.~Xu, and X.~Li, ``A novel center-boundary metric loss to learn discriminative features for hyperspectral image classification,'' \emph{IEEE Transactions on Geoscience and Remote Sensing}, vol.~62, pp. 1--16, 2024.

\bibitem{9565208}
Z.~Zhong, Y.~Li, L.~Ma, J.~Li, and W.-S. Zheng, ``Spectral–spatial transformer network for hyperspectral image classification: A factorized architecture search framework,'' \emph{IEEE Transactions on Geoscience and Remote Sensing}, vol.~60, pp. 1--15, 2022.

\bibitem{SSFTT}
L.~Sun, G.~Zhao, Y.~Zheng, and Z.~Wu, ``Spectral–spatial feature tokenization transformer for hyperspectral image classification,'' \emph{IEEE Transactions on Geoscience and Remote Sensing}, vol.~60, pp. 1--14, 2022.

\bibitem{zhu2024vision}
L.~Zhu, B.~Liao, Q.~Zhang, X.~Wang, W.~Liu, and X.~Wang, ``Vision mamba: Efficient visual representation learning with bidirectional state space model,'' \emph{arXiv preprint arXiv:2401.09417}, 2024.

\bibitem{wang2024graph}
C.~Wang, O.~Tsepa, J.~Ma, and B.~Wang, ``Graph-mamba: Towards long-range graph sequence modeling with selective state spaces,'' \emph{arXiv preprint arXiv:2402.00789}, 2024.

\bibitem{wang2023dcn}
D.~Wang, J.~Zhang, B.~Du, L.~Zhang, and D.~Tao, ``Dcn-t: Dual context network with transformer for hyperspectral image classification,'' \emph{IEEE Transactions on Image Processing}, vol.~32, pp. 2536--2551, 2023.

\bibitem{xie2026mhc}
\BIBentryALTinterwordspacing
Z.~Xie, Y.~Wei, H.~Cao, C.~Zhao, C.~Deng, J.~Li, D.~Dai, H.~Gao, J.~Chang, K.~Yu, L.~Zhao, S.~Zhou, Z.~Xu, Z.~Zhang, W.~Zeng, S.~Hu, Y.~Wang, J.~Yuan, L.~Wang, and W.~Liang, ``mhc: Manifold-constrained hyper-connections,'' 2026. [Online]. Available: \url{https://arxiv.org/abs/2512.24880}
\BIBentrySTDinterwordspacing

\bibitem{mishra2026mhc}
S.~Mishra, ``mhc-gnn: Manifold-constrained hyper-connections for graph neural networks,'' \emph{arXiv preprint arXiv:2601.02451}, 2026.

\bibitem{zhu2024hyper}
D.~Zhu, H.~Huang, Z.~Huang, Y.~Zeng, Y.~Mao, B.~Wu, Q.~Min, and X.~Zhou, ``Hyper-connections,'' \emph{arXiv preprint arXiv:2409.19606}, 2024.

\bibitem{hong2025hyperspectral}
D.~Hong, C.~Li, N.~Yokoya, B.~Zhang, X.~Jia, A.~Plaza, P.~Gamba, J.~A. Benediktsson, and J.~Chanussot, ``Hyperspectral imaging,'' \emph{arXiv preprint arXiv:2508.08107}, 2025.

\bibitem{dewis2025multitaskglocalobiamambasentinel2}
\BIBentryALTinterwordspacing
Z.~Dewis, Y.~Zhu, Z.~Xu, M.~Heffring, S.~Taleghanidoozdoozan, K.~Xiao, M.~Alkayid, and L.~L. Xu, ``Multitask glocal obia-mamba for sentinel-2 landcover mapping,'' 2025. [Online]. Available: \url{https://arxiv.org/abs/2511.10604}
\BIBentrySTDinterwordspacing

\bibitem{shi2022deep}
S.~Shi, L.~Zhang, Y.~Altmann, and J.~Chen, ``Deep generative model for spatial--spectral unmixing with multiple endmember priors,'' \emph{IEEE Transactions on Geoscience and Remote Sensing}, vol.~60, pp. 1--14, 2022.

\bibitem{ahmad2025graphmamba}
M.~Ahmad, M.~Mazzara, S.~Distefano, A.~M. Khan, M.~H.~F. Butt, M.~Usama, and D.~Hong, ``Graphmamba: Graph tokenization mamba for hyperspectral image classification,'' \emph{IEEE Transactions on Emerging Topics in Computing}, vol.~13, no.~4, pp. 1510--1521, 2025.

\end{thebibliography}

% that's all folks
\end{document}